\documentclass{article}
\usepackage[rgb]{xcolor}
\usepackage[final,nonatbib]{neurips_2023} 
\usepackage[sorting=none, style=numeric, bibencoding=utf8, backend=biber, natbib=true, uniquename=false, uniquelist=false, maxbibnames=99]{biblatex}
\usepackage{graphicx}
\usepackage{hyperref}
\usepackage{url}
\usepackage{colortbl} 
\usepackage{amsmath,amsfonts,amssymb}
\usepackage{lipsum}
\usepackage{multirow} 
\usepackage{bigdelim} 
\usepackage{hhline}
\usepackage{subcaption}
\usepackage{tcolorbox}
\usepackage{changepage}
\usepackage{sidecap}
\usepackage{paralist}
\usepackage{caption}
\usepackage{cleveref} 
\usepackage{svg}
\usepackage{enumitem}
\usepackage{graphicx}
\usepackage{amssymb}
\usepackage{wrapfig}
\usepackage{array}
\usepackage{soul}
\usepackage{tikz}
\usepackage[nonumberlist, toc]{glossaries}
\usepackage{pgfplots}
\pgfplotsset{compat=1.16}
\makeglossaries

\newcommand{\xclean}{x_{\text{clean}}}

\newcommand{\eclean}{e_{\text{clean}}}
\newcommand{\ecorr}{e_{\text{corr}}}
\definecolor{darkyellow}{RGB}{204, 204, 0}

\newtcbox{\mymath}[1][]{%
    nobeforeafter, math upper, tcbox raise base,
    enhanced, colframe=white!1!black,
    colback=white!1, boxrule=1pt,
    #1
}

\newcommand{\vocab}[1]{\textit{\textbf{#1}}}

\title{Attribution Patching Outperforms \\Automated Circuit Discovery}

\author{%
  Aaquib Syed \\
  University of Maryland, College Park \\
  \texttt{asyed04@umd.edu} \\
  \And
  Can Rager \\
  Independent \\
  \texttt{canrager@gmail.com} \\
  \And
  Arthur Conmy \\
  Independent \\
  \texttt{arthurconmy@gmail.com} \\
}

\begin{document}
\maketitle

\begin{abstract}
Automated interpretability research has recently attracted attention as a potential research direction that could scale explanations of neural network behavior to large models.
Existing automated circuit discovery work applies activation patching to identify subnetworks responsible for solving specific tasks (circuits).
In this work, we show that a simple method based on attribution patching outperforms all existing methods while requiring just two forward passes and a backward pass.
We apply a linear approximation to activation patching to estimate the importance of each edge in the computational subgraph. Using this approximation, we prune the least important edges of the network. We survey the performance and limitations of this method, finding that averaged over all tasks our method has greater AUC from circuit recovery than other methods.\footnote{Our code is available at \url{https://github.com/Aaquib111/acdcpp}}
\end{abstract}

\section{Introduction}
\label{sec:intro}

Mechanistic interpretability is a subfield of AI interpretability that focuses on attributing model behaviors to its components, thus reverse engineering the network \cite{olah2022interp}. This field aims to identify subnetworks (circuits) within the model which are responsible for solving specific tasks \cite{olah2020zoom}. Prior attempts at finding circuits in language models have led to finding networks of attention heads and multi-layer perceptrons (MLPs) that partially or fully explain model behaviors at tasks such as indirect object identification, modular arithmetic, completion of docstrings, and predicting successive dates \citep{wang2023interpretability, nanda2023progress, docstring, greaterthan}. However, almost all previous work has been limited to relatively small models since manually applying mechanistic interpretability methods has not currently scaled to end-to-end circuits in larger models \citep{lieberum2023does}.

It may be important to scale interpretability to large models as these are the neural networks most widely deployed and used by a wide range of people. Currently, we have little understanding into these models work and failure modes are not always found ahead of deployment. If successful, scaled interpretability could address a wide variety of concerns about the lack of transparency of language models \citep{vig2020causal}, in addition to speculative risks about the alignment of machine learning systems \citep{hubinger2020overview}.

Automated Circuit Discovery (ACDC; \citep{conmy2023automated}) attempts to automate a large portion of the mechanistic interpretability workflow --- the pruning of edges between attention heads and MLPs that do not affect the task being studied. ACDC begins with a computational graph, and recursively calculates the importance of an edge in the graph for a specific task. In our work, we use edges to refer to activations inside models between two components (\Cref{sec:related_work} describes this motivation further). ACDC's pruning algorithm applies \textbf{activation patching}. (Note that \textbf{activation patching} is not \textbf{attribution patching}. Both are defined in full in \Cref{sec:edge_attribution_patching}.) At a high level, activation patching edits a specific activation in a model forward pass and measures a model statistic (e.g loss) under this intervention. Activation patching is inefficient for circuit discovery because getting each statistic about model activations requires another forward pass. Our work uses \textbf{attribution patching} to recover circuits more efficiently (\Cref{sec:edge_attribution_patching}).

Our main contributions are:

\begin{compactenum}
    \item Introducing a method for using attribution patching on all computational graph edges for automated circuit discovery (Edge Attribution Patching, \Cref{sec:edge_attribution_patching}).
    \item Benchmarking Edge Attribution Patching vs existing circuit discovery methods (\Cref{sec:results}).
    \item Finding and explaining some limitations with Edge Attribution Patching (\Cref{sec:limitations}).
\end{compactenum}



\section{Related Work}
\label{sec:related_work}

\textbf{Automated Circuit Discovery} refers to finding the important subgraph of models' computational graphs for performance on particular tasks \citep{conmy2023automated}. Existing algorithms include efficient heuristics \citep{sixteen_heads} and gradient-descent based methods \citep{louizos2017learning, subnetwork_probing}. ACDC is related to pruning \citep{blalock2020state} and other compression techniques \citep{compression}, but differs in how the compressed networks are reflective of the circuits that model uses to compute outputs to certain tasks and the goal of ACDC is not to speed up forward passes (all techniques studied in this work slow forward passes). 

\textbf{Activation Patching} is a technique for analyzing the role of individual components in a model. It involves targeted manipulations of activations during a forward pass (further explained in \Cref{sec:activation_patching}). Previous works applied this technique under various names, such as Interchange Interventions \citep{causal_abstraction_geiger}, Causal Mediation Analysis \citep{vig2020causal} and Causal Tracing \citep{meng2022locating}. We adapt the terminology used by Conmy et al. \citep{conmy2023automated}.

\textbf{Transformer Circuits}. Our work builds upon the framework for understanding transformers for interpretability as introduced by \citet{elhage2021mathematical}. The important details include how they formulate forward passes of transformer models. Individual attention heads and MLPs (collectively called nodes) read and write information to a central communication channel, also called the residual stream. In these terms we can examine dependencies of nodes with the output of earlier nodes, i.e we can measure the effect of attention heads in layer 0 on the attention heads in layer 2. In the following, we view these dependencies as edges between nodes, building on existing work using this perspective \citep{docstring,greaterthan,wang2023interpretability}.

\section{Edge Attribution Patching}
\label{sec:edge_attribution_patching}

We present \textbf{Edge Attribution Patching} (EAP) as a technique to identify relevant model components for solving a specific task. In the following, we view language models as directed, acyclic graphs. In these terms, we aim to find small subgraphs that retain good performance on narrow tasks. We determine the importance of a specific edge through targeted manipulation of activations during a forward pass. We compare two approaches, Attribution Patching and Activation Patching, in order to motivate EAP.

\subsection{Activation Patching}
\label{sec:activation_patching}

\vocab{Activation patching} refers to replacing the activations from one model forward pass with the activations from a different forward pass. This method is typically applied to measure the counterfactual importance of model components, i.e. to measure a statistic $L(x)$ from model outputs under the activation patching, where $x$ is an input prompt. For example, $L$ often represents loss or logit difference \citep{wang2023interpretability}.

Following existing work (\Cref{sec:related_work}), we study the effect of activation patching on specific model edges by setting these equal to activations from different forward passes. Concretely, suppose that an edge $E$ in the computational graph has activation $e_\text{corr}$ on some corrupted prompt. In this work, we use the change in metric under activation patching 

\begin{equation}
    |L(\xclean |\ \text{do}(E = e_\text{corr})) - L(\xclean )|
    \label{eqn:activation_patching}
\end{equation}

to measure the impact of edge $E$. We use do-notation from causality \citep{pearl1995causal} to emphasise that activation patching is a causal intervention.

\subsection{Attribution Patching}
\label{sec:attribution_patching}

\begin{figure}
    \vspace*{-3.5cm}
    \centering

    \begin{subfigure}[b]{0.54\textwidth}
        \begin{tikzpicture}[font=\Large]
            \begin{axis}[
                view={45}{20},
                    axis lines=center,
                    ticks=none,
                    zlabel={$z$: $L$},
                ]
                
                \addplot3[surf, opacity=0.7, samples=20, domain=0:4, y domain=0:4] {4*x^2+4*y^2+5};
                
                \addplot3[surf, opacity=0.7, color=gray, samples=2, domain=0.5:4.5, y domain=0.5:4.5] {16*x+16*y-27};
                
                \addplot3[only marks, mark=*] coordinates {(2,2,37) (3.5,3.5,80)};
                
                \node[above] at (axis cs:2.0,1.7,20) {\(e_{\text{clean}}\)};
                \node[above] at (axis cs:4.0,3.9,57) {\(e_{\text{corr}}\)};
                \node[above] at (axis cs:4.0,1.0,0) {$(x, y)$\text{: Activation}};
                \draw[->, thick] (axis cs:2.01,2.01,37) -- (axis cs:3.4,3.4,80);
            \end{axis}
        \end{tikzpicture}
        \caption{Attribution Patching (\Cref{sec:edge_attribution_patching}) approximates the difference in metric $L$ caused by corrupting edges.}
        \label{fig:loss_landscape_lin_approx}
    \end{subfigure}
    \begin{subfigure}[b]{0.34\textwidth}
        \includegraphics[width=2.3\textwidth]{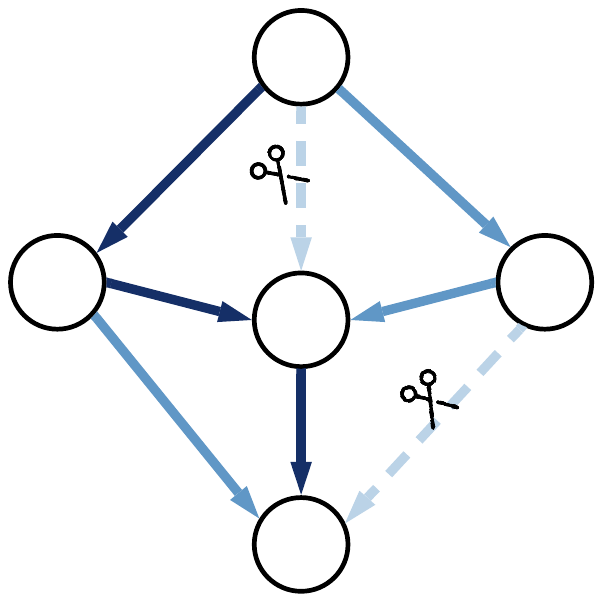}
        \caption{Removing the least important edges.}
        \label{fig:pruning_setp}
    \end{subfigure}
    
    \caption{Edge Attribution Patching (EAP)}
    \label{fig:eap_method}
\end{figure}

Activation patching slows ACDC since each measurement (like \Cref{eqn:activation_patching}) requires another forward pass. \vocab{Attribution patching} \citep{neelattribution} is a technique for estimating \Cref{eqn:activation_patching} for many different edges $E$ using only two forward passes and one backward pass.\footnote{Attribution patching (like activation patching) also applies to nodes and other model internal components that aren't edges, but we only use edges in this work.} It linearly approximates the metric difference after corrupting a single edge in the computational graph (\Cref{fig:eap_method}) by expanding $L$ as a function of the edge activation as a Taylor series with terms up to the first order: 

\begin{equation}
    L(\xclean |\ \text{do}(E=e_\text{corr})) \approx L(\xclean) + \underbrace{\left(e_{\text{corr}} - e_{\text{clean}}\right)^\top\frac{\partial}{\partial e_{\text{clean}}} L(\xclean |\ \text{do}(E=e_{\text{clean}}))}_{\text{Call this } \Delta_e L\text{, the \textbf{attribution score}.}}.
    \label{eqn:attribution_patching}
\end{equation}

A simple rearrangement implies that \Cref{eqn:activation_patching} is approximately equal to \hypertarget{eqn:abs_val}{$\left|\Delta_e L \right|$}~\hyperlink{eqn:abs_val}{(3)} which we call the \textbf{absolute attribution score} for the rest of this paper. In this work we always compute this score across a set of $(x_\text{clean}, x_\text{corr})$ pairs and take the mean.

In practice, all gradients needed to calculate the attribution scores come from intermediate terms computed in one ordinary backwards pass\footnote{In \Cref{app:only_one_backwards_pass} we show how only one backwards pass is required.} in PyTorch \citep{pytorch}, hence attribution patching is extremely efficient.

\subsection{Edge Attribution Patching}
\label{sec:edge_attribution_patching}

We can use the insights from \Cref{sec:attribution_patching} to build an automated circuit discovery algorithm. This takes two steps: i) use Equation \hyperlink{eqn:abs_val}{(2)} to obtain absolute attribution scores for the importance of all edges in the computational graph and then ii) sort these scores and keep the top $k$ edges in a circuit.
We use \textbf{Edge Attribution Patching} (EAP) to refer to this algorithm.
In the rest of the work we report results for all $k$ values when we evaluate EAP (similar to HISP in \citep{conmy2023automated}).


Note that one limitation of attribution patching is that it will not work when the gradient of the metric is the zero vector. \citet{conmy2023automated} recommended the use of KL divergence as a metric, which is i) equal to 0 when we run the model without ablations and ii) a non-negative metric. Therefore the zero point is a global minimum and hence all gradients are the zero vector at this point. In this work we use the `task-specific metrics' (not KL divergence) from \citep{conmy2023automated} so avoid this issue.

\section{Results}
\label{sec:results}
\subsection{Edge Attribution Patching vs Activation Patching vs ACDC}

We compare Edge Attribution Patching (EAP) and ACDC on the Indirect Object Identification (IOI), Docstring, and Greater-Than tasks. For each of these tasks, previous studies identified a subgraph (circuit) relevant for solving the task. We use their results as a ground truth for benchmarking both methods. We also compare using ACDC with the task-specific metrics (e.g logit difference) and KL Divergence (which was originally recommended). For the docstring task, we also include repeated activation patching as another point of reference for performance comparisons. We applied repeated activation patching by running the same circuit discovery method described in \Cref{sec:edge_attribution_patching} but using \Cref{eqn:activation_patching} rather than absolute attribution scores. Activation patching was not included in the other tasks as it was too computationally expensive to run on the GPT-2 small models used by IOI and Greater-Than. Subnetworks found using EAP for all three tasks are shown in \Cref{sec:eap_subnetworks}. 

\definecolor{darkpink}{RGB}{204, 0, 102} 

\begin{figure}[!h]
  \centering
  \begin{subfigure}[b]{0.33\textwidth}
        \includegraphics[width=\textwidth]{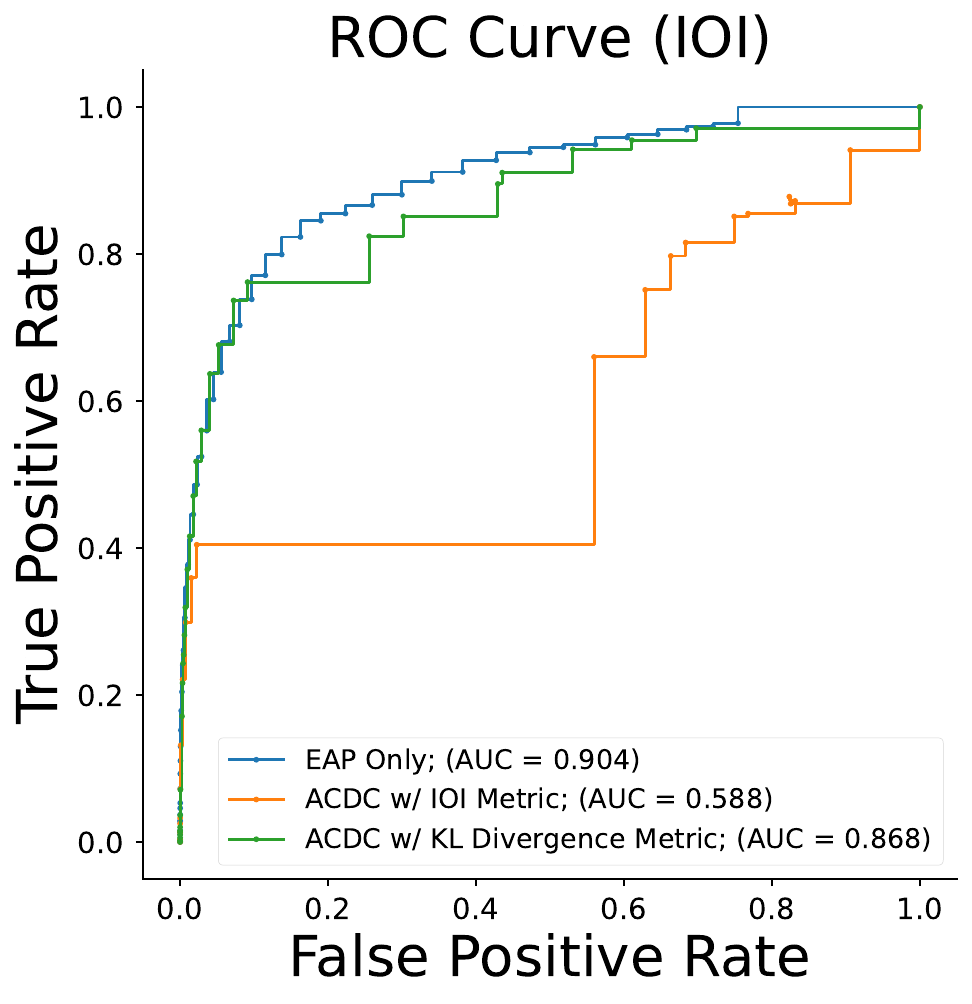}
    \caption{IOI task}
    \label{fig:ioi_roc}
  \end{subfigure}
  \hfill
  \begin{subfigure}[b]{0.32\textwidth}
    \includegraphics[width=\textwidth]{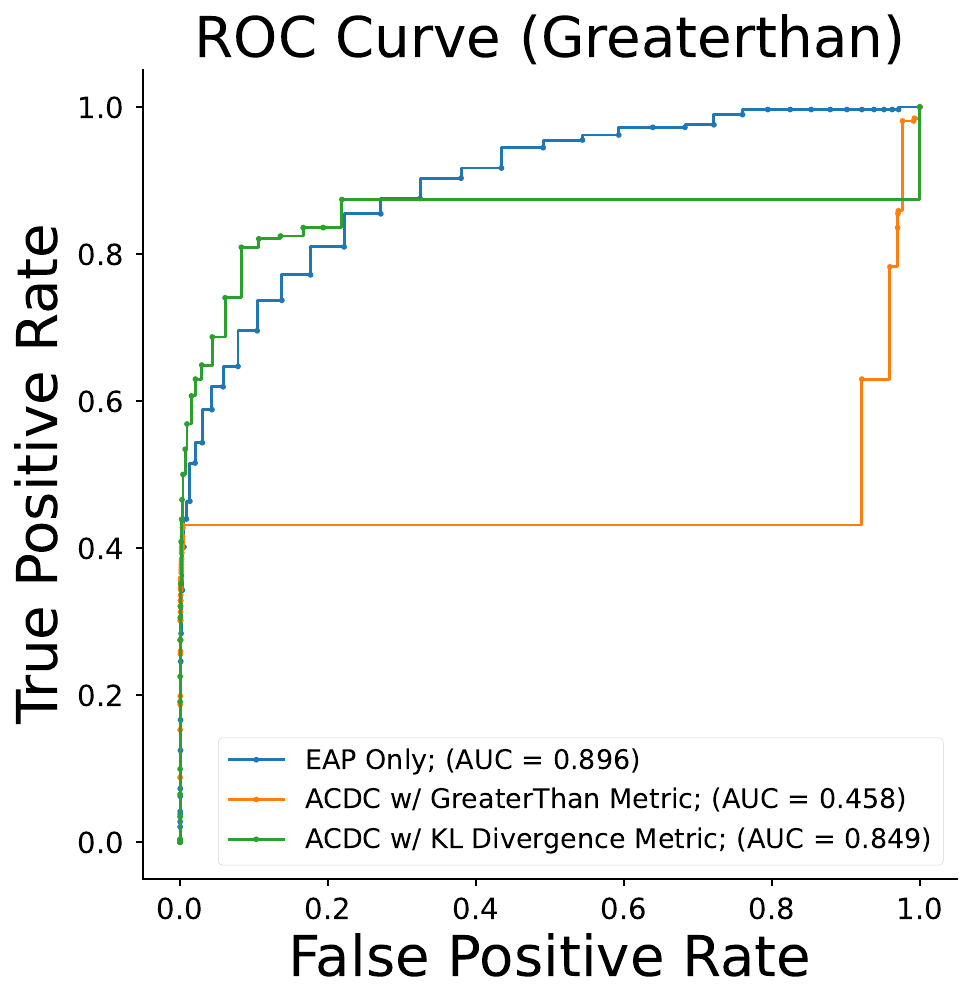}
    \caption{Greater-Than task}
    \label{fig:greaterthan_roc}
  \end{subfigure}
  \hfill
  \begin{subfigure}[b]{0.33\textwidth}
    \includegraphics[width=1.05\textwidth]{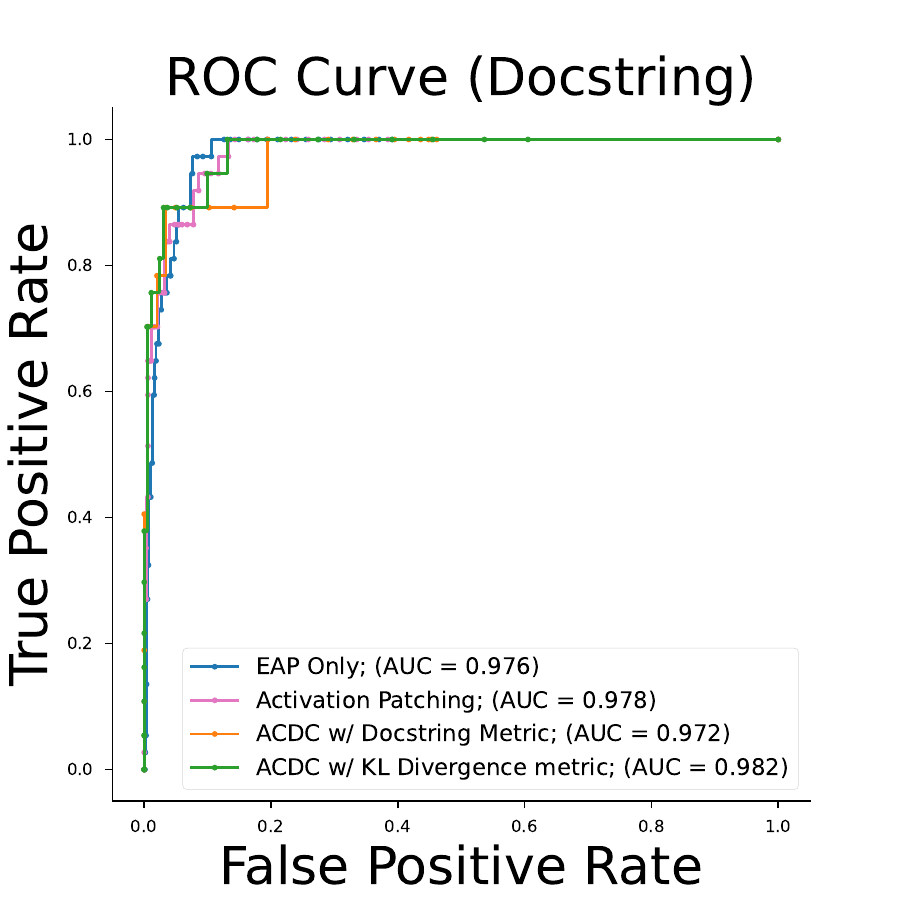}
    \caption{Docstring task}
    \label{fig:docstring_roc}
  \end{subfigure}
  
  
    \caption{ROC Curves comparing \textcolor{blue}{EAP}, \textcolor{orange}{ACDC with task metric}, and \textcolor{green}{ACDC with KL Divergence}. The Docstring plot also compares to \textcolor{darkpink}{Activation Patching}.}
  \label{fig:roc_curves}
\end{figure}

The ROC curves in Figure \ref{fig:roc_curves} suggest the performance of EAP is better than ACDC overall: it has the maximal AUC in \Cref{fig:ioi_roc}-\ref{fig:greaterthan_roc}, while ACDC used with the KL Divergence metric outperforms EAP in \Cref{fig:docstring_roc}. 
ACDC outperformed the existing methods HISP and Subnetwork Probing methods \citep{conmy2023automated}. We conclude EAP outperforms all previous methods for circuit discovery, since it is competitive with ACDC on recovering circuits while significantly reducing the computational demand: EAP only takes a constant number of forward and backwards passes while the number of forward passes required by ACDC is scaling exponentially with the number of nodes. 

\subsection{Validating EAP Attribution Scores}
In this section, we look at the approximate metric change (attribution score) EAP assigns to each edge in the model. We aim to understand the relation between the attribution score and the function of the edge in the task being studied. First, we look at the distribution of scores for edges in the circuit compared to edges not in the circuit for each of the three tasks.

\begin{figure}[!hbt]
    \centering
    \includegraphics[width=0.5\textwidth]{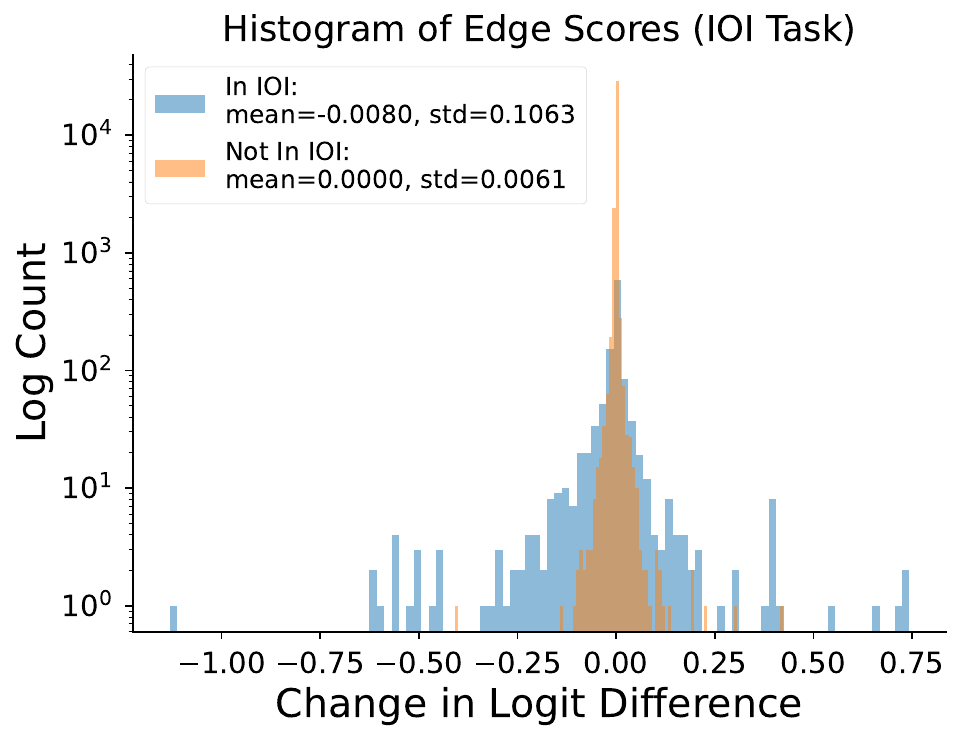}
    \caption{Distribution of Attribution Scores for the IOI Task (Logit Diff)}
    \label{fig:attr_dist_ioi}
\end{figure}

Figure \ref{fig:attr_dist_ioi} shows the distribution of attribution scores for the IOI task. The distributions for the remaining tasks can be found in \Cref{sec:score_dists}. Qualitatively, attribution scores for edges in the circuit tend to be spread further from zero. Furthermore, there are only 6 edges outside of the interval $[-0.25, 0.25]$ that aren't part of the IOI circuit. 
We further explore the attribution scores for the IOI circuit's classes of heads in \Cref{app:edge_roles}.

\section{Limitations}
\label{sec:limitations}



We introduced edge activation patching as an approximation to activation patching. However, we found that edge activation patching outperformed ACDC, a technique based on activation patching (\Cref{sec:results}). In this section, we investigate whether attribution patching's success is due to extremely accurate approximations (in \Cref{subsec:attrib_activ_approx} we find that the answer is no), and whether there is any further use for ACDC (in \Cref{subsec:acdc_use} we find that the answer is yes). We use the docstring task as a case study due to the small model size used.

\subsection{How faithful are Attribution Patching's approximations?}
\label{subsec:attrib_activ_approx}
\begin{figure}[htb!]
    \centering
    \begin{subfigure}[t]{0.48\textwidth}
        \vspace{0pt}
        \includegraphics[width=\textwidth]{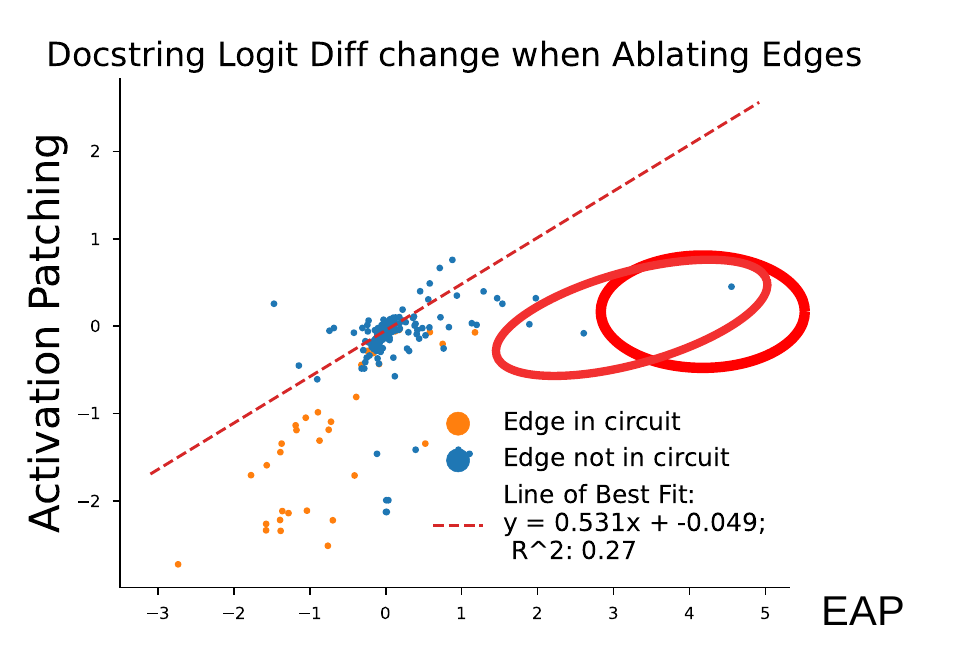}
        \caption{Distribution of attribution scores for edges from activation patching and attribution patching. Circled: outlier EAP point studied in \Cref{fig:study-linear-approx}.}
        \label{fig:scatter}
    \end{subfigure}
    \hfill
    \begin{subfigure}[t]{0.48\textwidth}
        \vspace{0.4cm}
        \includegraphics[width=\textwidth]{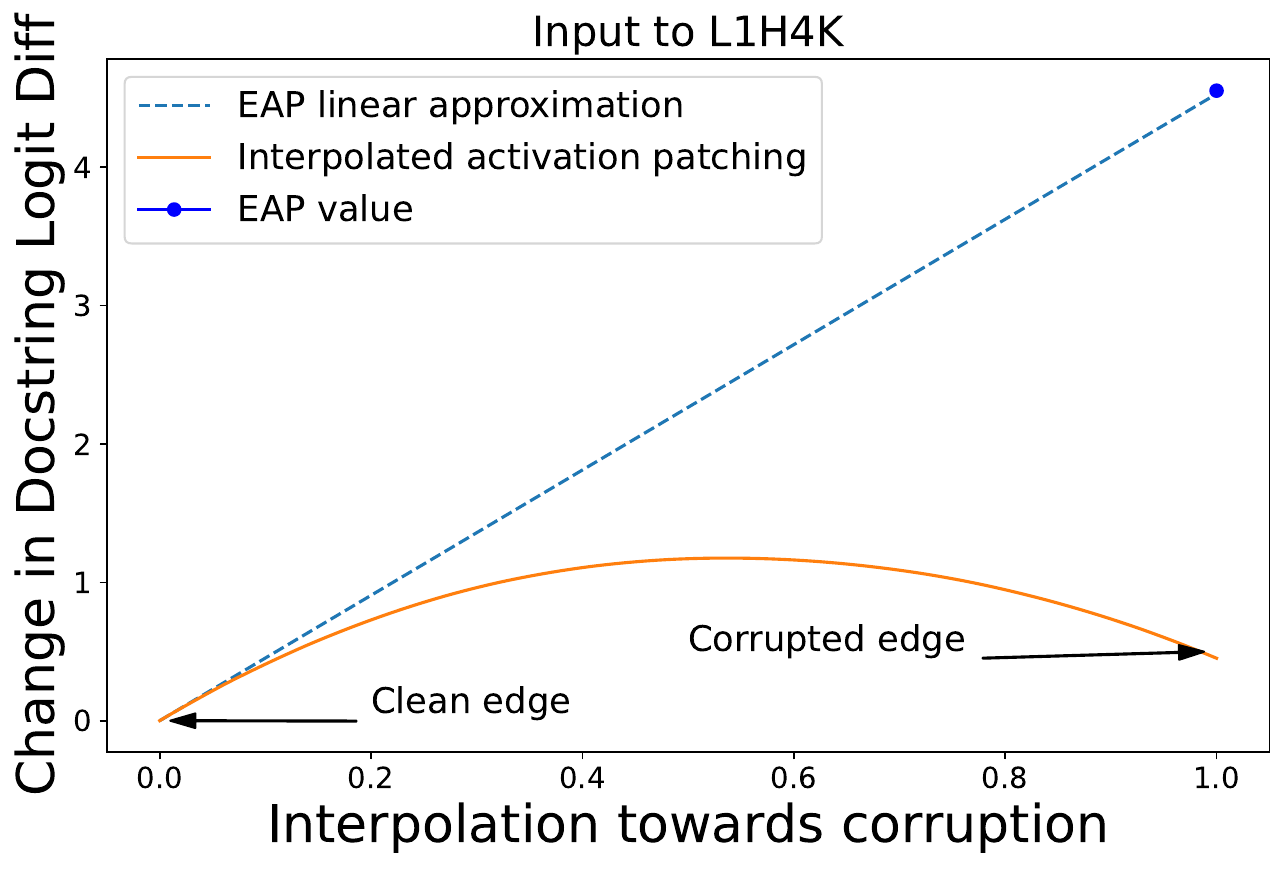}
        \caption{Visualizing the rightmost point in \Cref{fig:scatter}. Note that corrupting this edge (surprisingly) slightly increases the logit difference on the Docstring task (higher logit difference is better). However, EAP overestimates how large this increase is.}
        \label{fig:study-linear-approx}
    \end{subfigure}
    \caption{Visualizing Edge Attribution Patching.}
\end{figure}

To study how faithful the approximation \Cref{eqn:attribution_patching} is, we plot the attribution patching scores (Equation \hyperlink{eqn:abs_val}{(2)}) against the activation patching scores (\Cref{eqn:activation_patching}) in \Cref{fig:scatter}. Surprisingly, we find a fairly weak correlation between activation and attribution patching scores ($R^2 = 0.27$). Further, the line of best fit has gradient 0.531, suggesting that attribution patching estimates the effect of activation patching as twice as important as it really is. 

Moreover, we can gain some sense for the discrepancy between activation and attribution patching by studying the continuous transition between clean ($\eclean$) and corrupted ($\ecorr$) activations in \Cref{eqn:activation_patching}, i.e studying the values $|L(\xclean |\ \text{do}(E = \lambda e_\text{corr} + (1 - \lambda) \eclean )) - L(\xclean )|$ for $0 \le \lambda \le 1$. We can compare this to the linear approximations of Attribution Patching $\lambda \Delta_e L$. \Cref{fig:study-linear-approx} shows the result for one edge in the docstring circuit where the linear approximation to activation patching is not accurate.

We find that interpolating towards the corrupted input creates a concave curve (\Cref{fig:study-linear-approx}) such that the linear approximation at $\lambda = 0$ overestimates the effect of activation patching this edge. In \Cref{app:more_quadratic} we show that this also holds for the other outlier edges in the ellipse in \Cref{fig:scatter}.

\subsection{Is there any further use for ACDC?}
\label{subsec:acdc_use}

\begin{figure}[!h]
    \centering
    \includegraphics[width=0.7\textwidth]{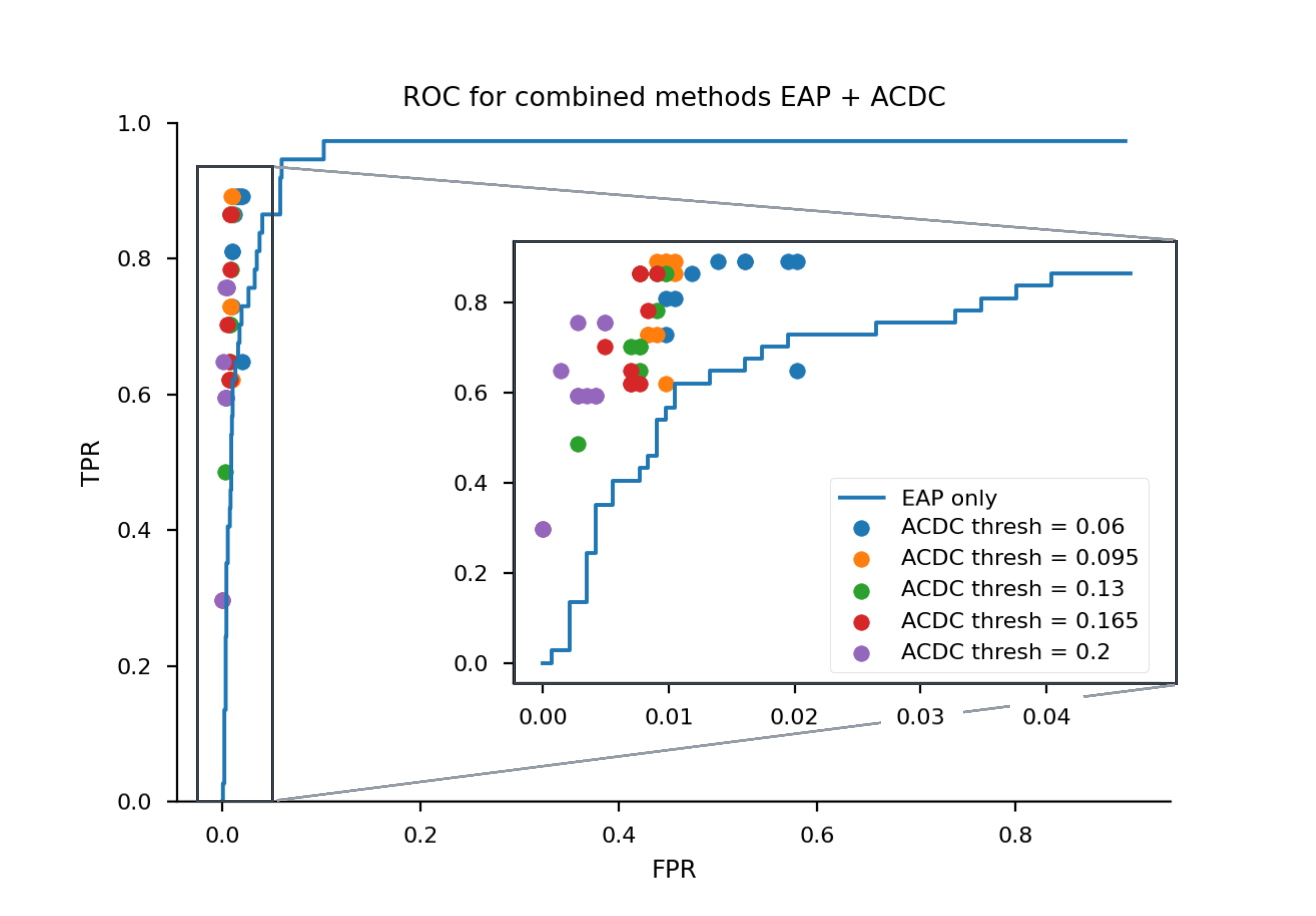}
    \caption{Comparing statistics of the combined EAP + ACDC methods with EAP only. The inset shows a zoom to the significant area of the statistics of the combined method.}
    \label{fig:combined_roc}
\end{figure}

In Section \ref{subsec:attrib_activ_approx} above, we found that EAP overestimates activation patching in cases where the attribution score is concave. This suggests the potential to refine the result by running ACDC on the pruned subgraph returned by EAP. We ran EAP first, then ACDC on the resulting subgraph for the Docstring task, varying pruning thresholds for EAP and ACDC independently. Figure \ref{fig:combined_roc} compares the TPR and FPR for the combined methods with the ROC curve of EAP only. The combined methods show increased performance compared to EAP only. 

Finally, one further limitation of this research is that the metrics used for interpretability do not precisely capture meaningful human understanding. Recovering a subgraph that humans previously recovered is limited because i) we can't evaluate this metric for interpretability tasks that we don't yet understand and ii) human-found circuits are imperfect, increasing the noise in this measurement.

\section{Conclusion}

We provide evidence that Edge Attribution Patching (EAP) outperforms ACDC in identifying circuits while being substantially faster to compute. This result is surprising, as EAP is an approximation for activation patching, the method applied by ACDC. However, running ACDC on the prepruned subnetwork found by EAP can improve the identification of relevant edges. Therefore, we suggest future circuit discovery experiments to run EAP first and then apply ACDC.

\ifunderreview{}{
    \section{Author Contributions}
    Aaquib Syed and Can Rager proposed combining ACDC with attribution patching methods and implemented initial prototypes. Arthur Conmy advised working on attributing edges rather than nodes and Aaquib made the first findings that this outperformed Automatic Circuit Discovery. All authors worked on the paper's figures, experiments and code.
    \section{Acknowledgements}
    We would like to thank Callum McDougall for organising ARENA and providing a great introduction to interpretability, and Rusheb Shah and Lucy Farnik for collaboration on the ARENA hackathon prototype which this work is based on.
    We would like to thank Neel Nanda for a helpful discussion and János Kramár, Stephen Casper and Euan Ong for suggestions based on a earlier version of this work. 
}
\newpage

\printbibliography

@inproceedings{
wang2023interpretability,
title={Interpretability in the Wild: a Circuit for Indirect Object Identification in {GPT}-2 Small},
author={Kevin Ro Wang and Alexandre Variengien and Arthur Conmy and Buck Shlegeris and Jacob Steinhardt},
booktitle={The Eleventh International Conference on Learning Representations },
year={2023},
url={https://openreview.net/forum?id=NpsVSN6o4ul}
}

@inproceedings{meng2022locating,
  author    = {Meng, Kevin and Bau, David and Andonian, Alex J and Belinkov, Yonatan},
  booktitle = {Advances in Neural Information Processing Systems},
  year      = {2022},
  title     = {Locating and editing factual associations in {GPT}},
}

@article{olah2020zoom,
  author = {Olah, Chris and Cammarata, Nick and Schubert, Ludwig and Goh, Gabriel and Petrov, Michael and Carter, Shan},
  title = {Zoom In: An Introduction to Circuits},
  journal = {Distill},
  year = {2020},
  doi = {10.23915/distill.00024.001}
}

@online{olah2022interp,
  author = {Olah, Chris},
  year   = {2022},
  title  = {Mechanistic Interpretability, Variables, and the Importance of Interpretable Bases},
  url = {https://www.transformer-circuits.pub/2022/mech-interp-essay},
}

@inproceedings{subnetwork_probing,
  author    = {Cao, Steven and Sanh, Victor and Rush, Alexander},
  publisher = {Association for Computational Linguistics},
  booktitle = {Proceedings of the 2021 Conference of the North American Chapter of the Association for Computational Linguistics: Human Language Technologies},
  year      = {2021},
  doi       = {10.18653/v1/2021.naacl-main.74},
  pages     = {960--966},
  title     = {Low-Complexity Probing via Finding Subnetworks},
}

@article{elhage2021mathematical,
   title={A Mathematical Framework for Transformer Circuits},
   author={Elhage, Nelson and Nanda, Neel and Olsson, Catherine and Henighan, Tom and Joseph, Nicholas and Mann, Ben and Askell, Amanda and Bai, Yuntao and Chen, Anna and Conerly, Tom and DasSarma, Nova and Drain, Dawn and Ganguli, Deep and Hatfield-Dodds, Zac and Hernandez, Danny and Jones, Andy and Kernion, Jackson and Lovitt, Liane and Ndousse, Kamal and Amodei, Dario and Brown, Tom and Clark, Jack and Kaplan, Jared and McCandlish, Sam and Olah, Chris},
   year={2021},
   journal={Transformer Circuits Thread},
   url={https://transformer-circuits.pub/2021/framework/index.html}
}

@misc{greaterthan,
      title={How does GPT-2 compute greater-than?: Interpreting mathematical abilities in a pre-trained language model}, 
      author={Michael Hanna and Ollie Liu and Alexandre Variengien},
      year={2023},
      eprint={2305.00586},
      archivePrefix={arXiv},
      primaryClass={cs.CL}
}

@inproceedings{nanda2023progress,
  author    = {Nanda, Neel and Chan, Lawrence and Lieberum, Tom and Smith, Jess and Steinhardt, Jacob},
  url       = {https://openreview.net/forum?id=9XFSbDPmdW},
  booktitle = {The Eleventh International Conference on Learning Representations},
  year      = {2023},
  title     = {Progress measures for grokking via mechanistic interpretability},
}

@inproceedings{sixteen_heads,
  author    = {Michel, Paul and Levy, Omer and Neubig, Graham},
  editor    = {Wallach, Hanna M. and Larochelle, Hugo and Beygelzimer, Alina and d'Alch\'{e}{-}Buc, Florence and Fox, Emily B. and Garnett, Roman},
  url       = {https://proceedings.neurips.cc/paper/2019/hash/2c601ad9d2ff9bc8b282670cdd54f69f-Abstract.html},
  booktitle = {Advances in Neural Information Processing Systems 32: Annual Conference on Neural Information Processing Systems 2019, NeurIPS 2019, December 8-14, 2019, Vancouver, BC, Canada},
  year      = {2019},
  pages     = {14014--14024},
  title     = {Are Sixteen Heads Really Better than One?},
}

@misc{causal_abstraction_geiger,
  author    = {Geiger, Atticus and Lu, Hanson and Icard, Thomas and Potts, Christopher},
  publisher = {arXiv},
  url       = {https://arxiv.org/abs/2106.02997},
  year      = {2021},
  title     = {Causal Abstractions of Neural Networks},
}

@misc{docstring,
  author = {Heimersheim, Stefan and Janiak, Jett},
  url    = {https://www.alignmentforum.org/posts/u6KXXmKFbXfWzoAXn/a-circuit-for-python-docstrings-in-a-4-layer-attention-only},
  year   = {2023},
  title  = {A circuit for {P}ython docstrings in a 4-layer attention-only transformer},
}

@misc{lieberum2023does,
      title={Does Circuit Analysis Interpretability Scale? Evidence from Multiple Choice Capabilities in Chinchilla}, 
      author={Tom Lieberum and Matthew Rahtz and János Kramár and Neel Nanda and Geoffrey Irving and Rohin Shah and Vladimir Mikulik},
      year={2023},
      eprint={2307.09458},
      archivePrefix={arXiv},
      primaryClass={cs.LG}
}

@inproceedings{blalock2020state,
  author    = {Blalock, Davis W. and Ortiz, Jose Javier Gonzalez and Frankle, Jonathan and Guttag, John V.},
  editor    = {Dhillon, Inderjit S. and Papailiopoulos, Dimitris S. and Sze, Vivienne},
  publisher = {mlsys.org},
  url       = {https://proceedings.mlsys.org/book/296.pdf},
  booktitle = {Proceedings of Machine Learning and Systems 2020, MLSys 2020, Austin, TX, USA, March 2-4, 2020},
  year      = {2020},
  title     = {What is the State of Neural Network Pruning?},
}

@inproceedings{louizos2017learning,
  author    = {Louizos, Christos and Welling, Max and Kingma, Diederik P.},
  publisher = {OpenReview.net},
  url       = {https://openreview.net/forum?id=H1Y8hhg0b},
  booktitle = {6th International Conference on Learning Representations, {ICLR} 2018, Vancouver, BC, Canada, April 30 - May 3, 2018, Conference Track Proceedings},
  year      = {2018},
  title     = {Learning Sparse Neural Networks through $L_0$ Regularization},
}

@misc{neelattribution,
    author = {Nanda, Neel},
    title = {Attribution Patching: Activation Patching At Industrial Scale},
    url = {https://www.neelnanda.io/mechanistic-interpretability/attribution-patching},
    year = {2023}
}

@misc{hubinger2020overview,
      title={An overview of 11 proposals for building safe advanced AI}, 
      author={Evan Hubinger},
      year={2020},
      eprint={2012.07532},
      archivePrefix={arXiv},
      primaryClass={cs.LG}
}

@misc{vig2020causal,
      title={Causal Mediation Analysis for Interpreting Neural NLP: The Case of Gender Bias}, 
      author={Jesse Vig and Sebastian Gehrmann and Yonatan Belinkov and Sharon Qian and Daniel Nevo and Simas Sakenis and Jason Huang and Yaron Singer and Stuart Shieber},
      year={2020},
      eprint={2004.12265},
      archivePrefix={arXiv},
      primaryClass={cs.CL}
}

@misc{compression,
      title={A Survey on Model Compression for Large Language Models}, 
      author={Xunyu Zhu and Jian Li and Yong Liu and Can Ma and Weiping Wang},
      year={2023},
      eprint={2308.07633},
      archivePrefix={arXiv},
      primaryClass={cs.CL}
}

@inproceedings{conmy2023automated,
      title={Towards Automated Circuit Discovery for Mechanistic Interpretability}, 
      author={Arthur Conmy and Augustine N. Mavor-Parker and Aengus Lynch and Stefan Heimersheim and Adri{\`a} Garriga-Alonso},
      booktitle={Thirty-seventh Conference on Neural Information Processing Systems},
      year={2023},
      eprint={2304.14997},
      archivePrefix={arXiv},
      primaryClass={cs.LG}
}

@incollection{pytorch,
    title = {PyTorch: An Imperative Style, High-Performance Deep Learning Library},
    author = {Paszke, Adam and Gross, Sam and Massa, Francisco and Lerer, Adam and Bradbury, James and Chanan, Gregory and Killeen, Trevor and Lin, Zeming and Gimelshein, Natalia and Antiga, Luca and Desmaison, Alban and Kopf, Andreas and Yang, Edward and DeVito, Zachary and Raison, Martin and Tejani, Alykhan and Chilamkurthy, Sasank and Steiner, Benoit and Fang, Lu and Bai, Junjie and Chintala, Soumith},
    booktitle = {Advances in Neural Information Processing Systems 32},
    pages = {8024--8035},
    year = {2019},
    publisher = {Curran Associates, Inc.},
    url = {http://papers.neurips.cc/paper/9015-pytorch-an-imperative-style-high-performance-deep-learning-library.pdf}
}

@article{pearl1995causal,
  title={Causal diagrams for empirical research},
  author={Pearl, Judea},
  journal={Biometrika},
  volume={82},
  number={4},
  pages={669--688},
  year={1995},
  publisher={Oxford University Press}
}
\newpage
\appendix

\section{EAP Subnetworks}
\label{sec:eap_subnetworks}

\begin{figure}[!h]
\centering
  \begin{subfigure}[b]{\textwidth}
    \centering
    \includegraphics[width=\textwidth]{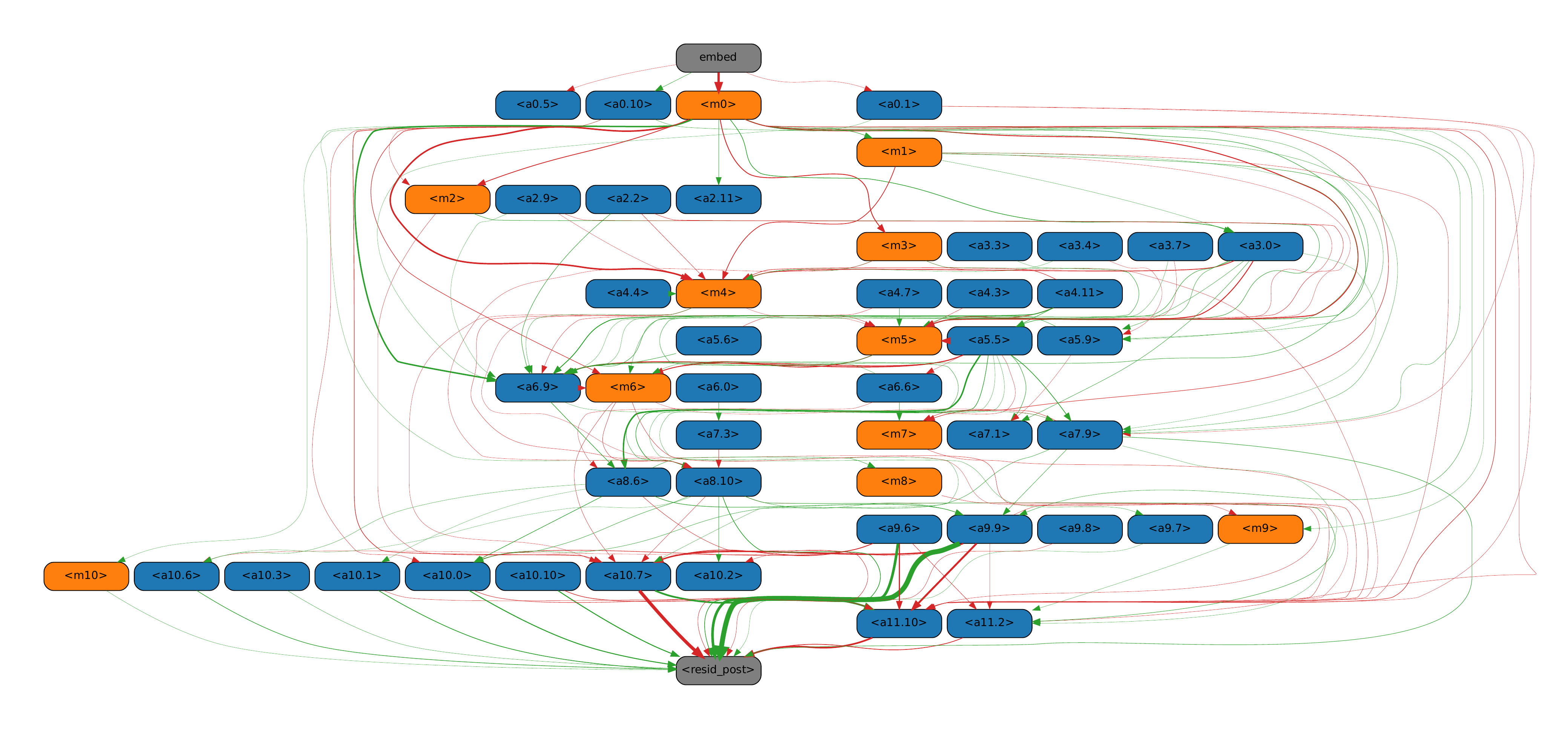}
    \caption{IOI Subnetwork, Threshold=0.077}
    \label{fig:ioi_subnetwork}
  \end{subfigure}
  \centering
  \begin{subfigure}[b]{0.8\textwidth}
    \centering
    \includegraphics[width=\textwidth]{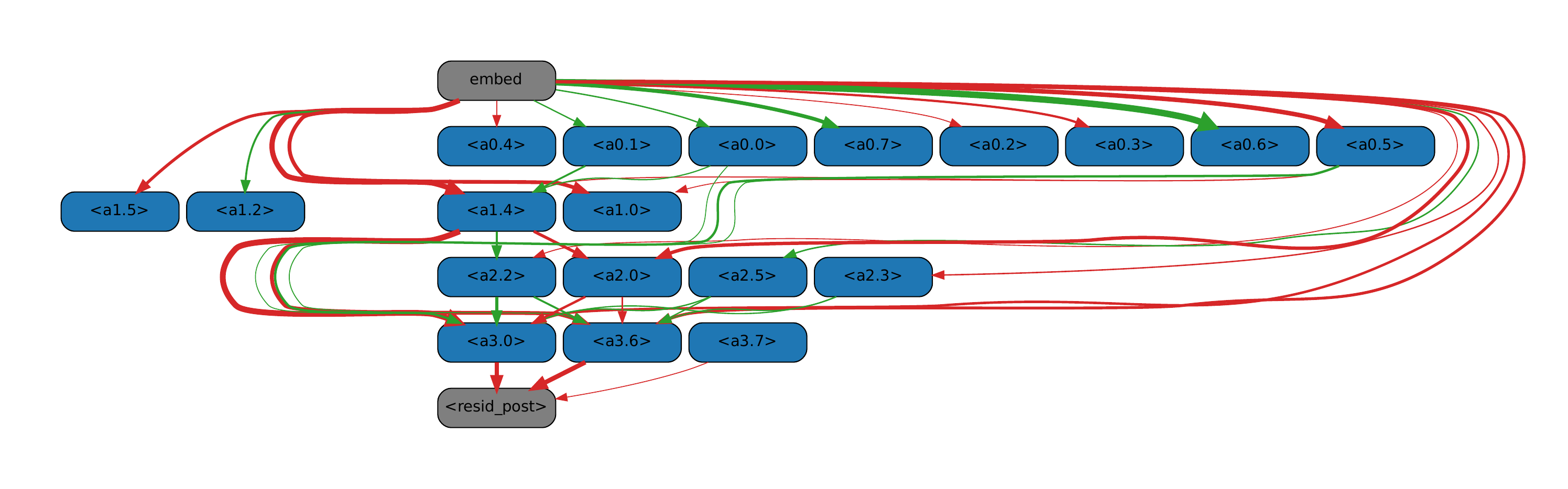}
    \caption{Docstring Subnetwork, Threshold=0.244}
    \label{fig:docstring_subnetwork}
  \end{subfigure}
  \centering
  \begin{subfigure}[b]{\textwidth}
    \centering
    \includegraphics[width=0.8\textwidth]{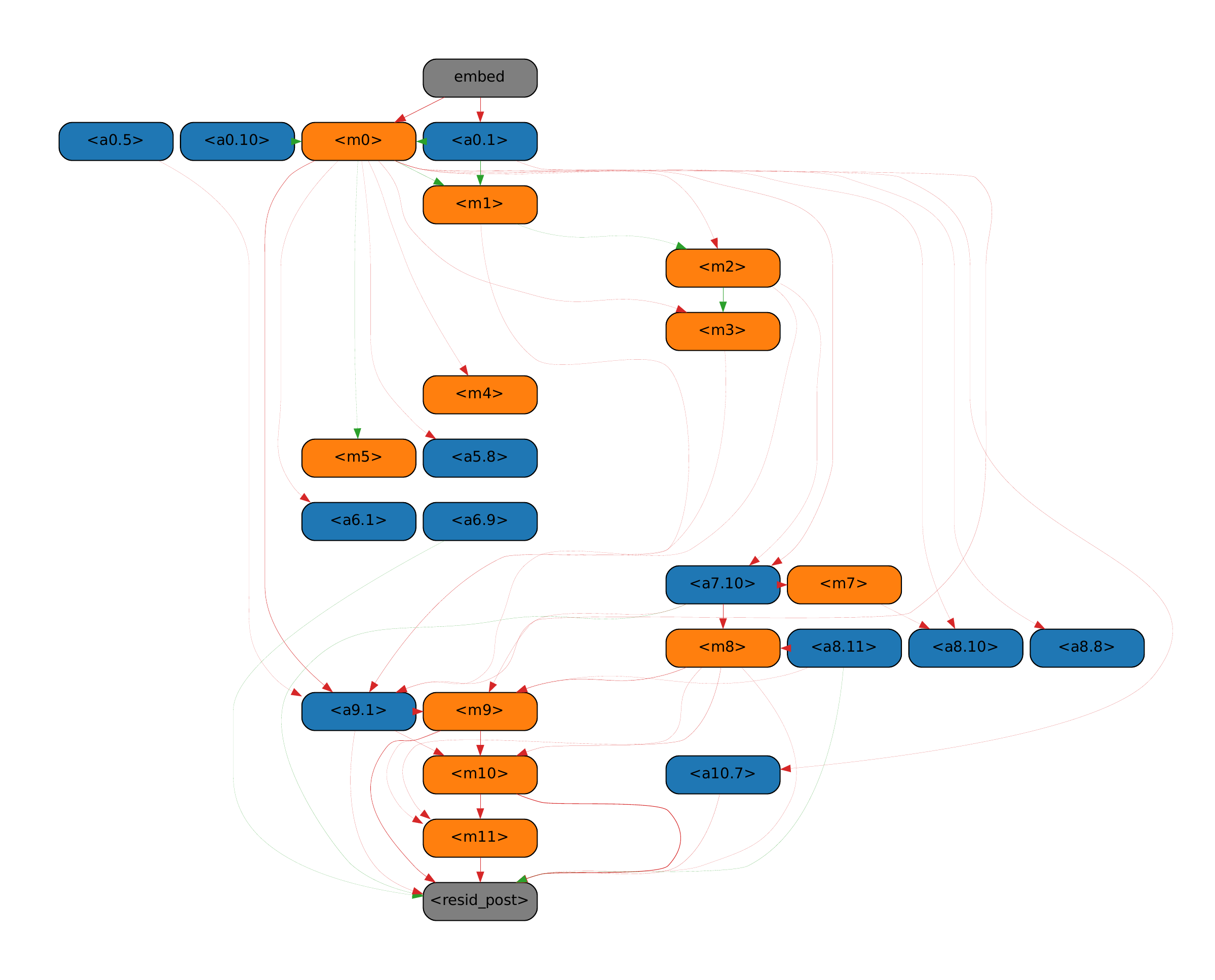}
    \caption{Greaterthan Subnetwork, Threshold=0.009}
    \label{fig:greaterthan_subnetwork}
  \end{subfigure}
  \caption{Resulting subnetworks after EAP at the given thresholds.}
  \label{fig:subnetworks}
\end{figure}

\newpage
\section{Distribution of EAP Attribution Scores}
\label{sec:score_dists}
\begin{figure}[!h]
  \begin{subfigure}[b]{0.49\textwidth}
    \includegraphics[width=\textwidth]{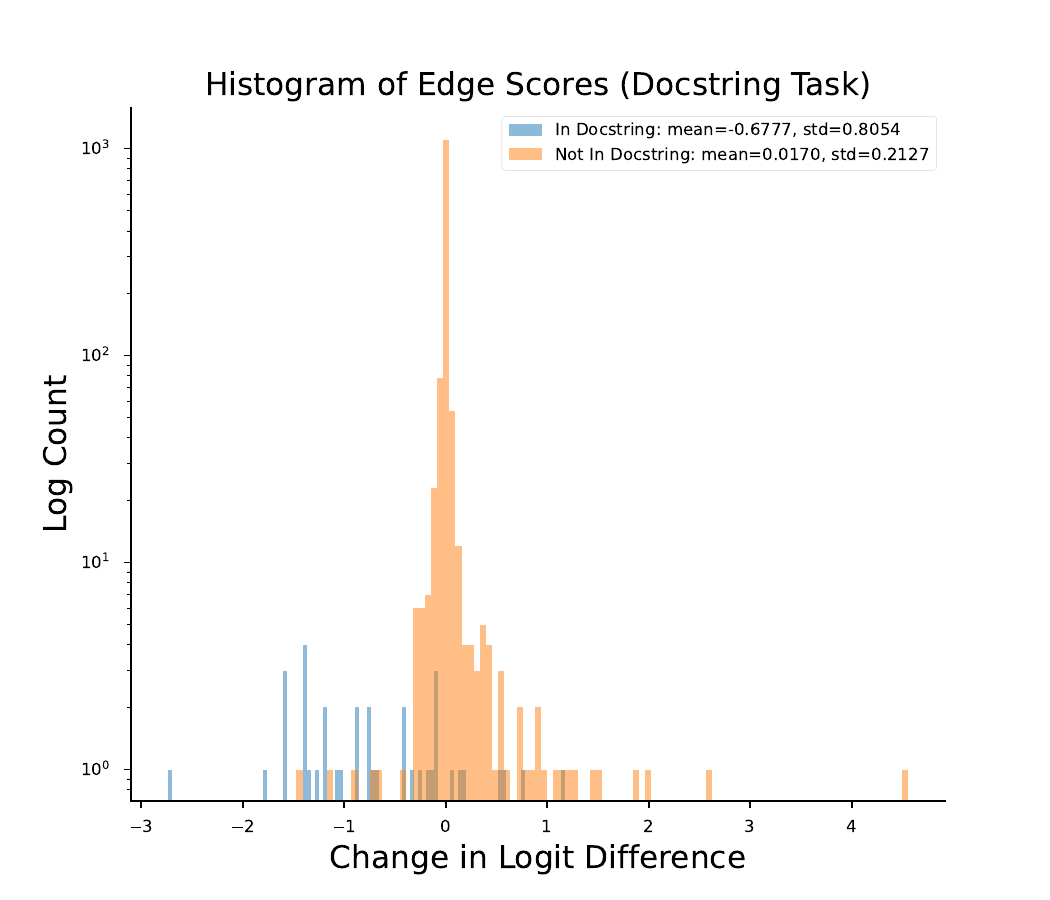}
    \caption{Distribution of Attribution Scores for the Docstring Task}
    \label{fig:docstring_hist}
  \end{subfigure}
  \hfill
  \begin{subfigure}[b]{0.49\textwidth}
    \includegraphics[width=\textwidth]{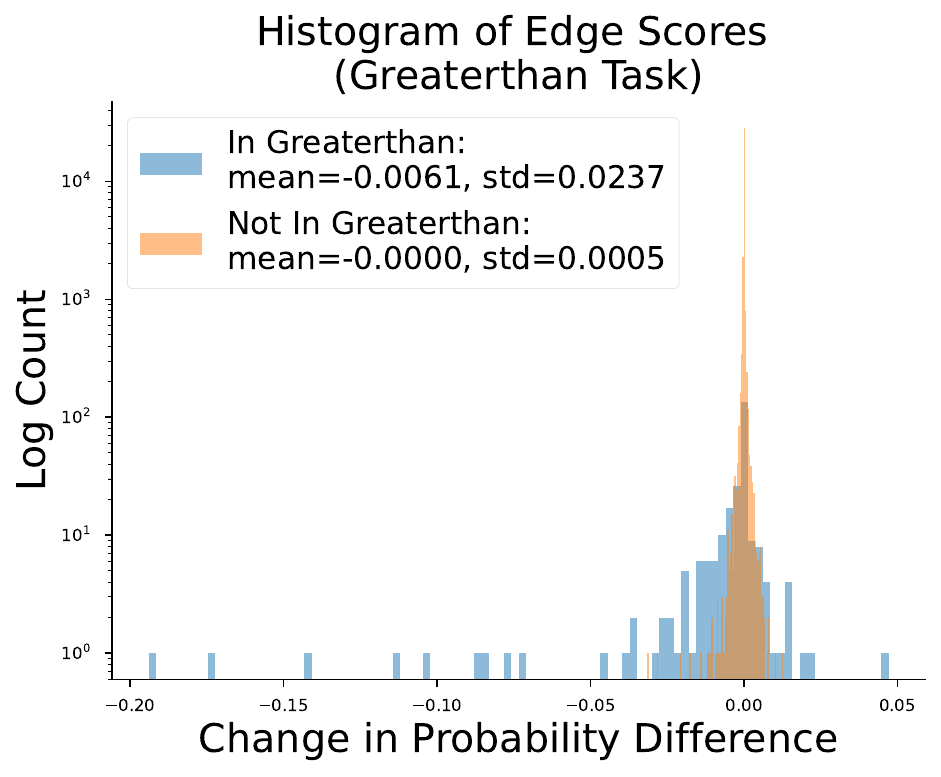}
    \caption{Distribution of Attribution Scores for the Greater-Than Task}
    \label{fig:greaterthan_hist}
  \end{subfigure}
  \caption{Distribution of Attribution Scores for the Docstring and Greater-Than tasks}
  \label{fig:score_dists}
\end{figure}

\section{Further investigation into combining EAP with ACDC}
\label{app:combined_methods}

\begin{figure}[htb!]
  \centering
  \includegraphics[width=\textwidth]{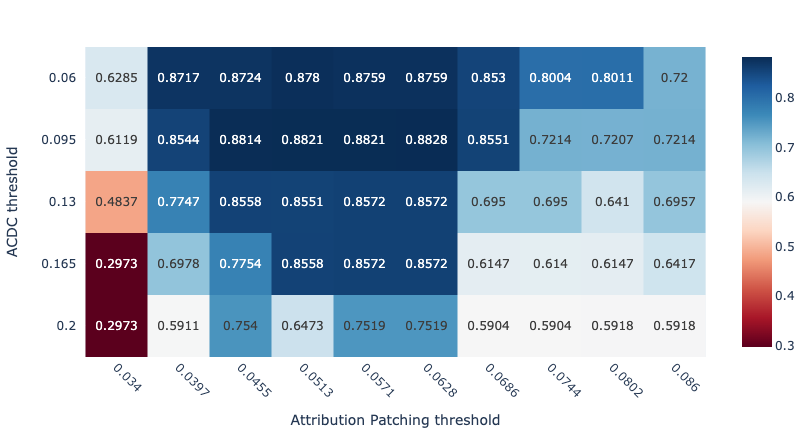}
  \caption{Youdens-J statistic (maximum TPR minus FPR value) for combining EAP and ACDC methods on the docstring task. We applied ACDC to the pruned subgraph returned by EAP.}
  \label{fig:thresh_sweep}
\end{figure}







\section{Further failures of attribution patching approximation}
\label{app:more_quadratic}

In \Cref{fig:my-app} we show further cases where in the docstring task attribution patching can be misleading. These cases all involve an edge that comes from the model's embeddings (positional and tokens). Our interpretation is that weighted averages of embeddings are anomalous inputs to the model and cause the concave change in docstring logit diff which doesn't occur when edges ae between non-embedding model components.

\begin{figure}[!bht]
    \centering
    \begin{subfigure}[t]{0.47\textwidth}
        \centering
        \includegraphics[width=\textwidth]{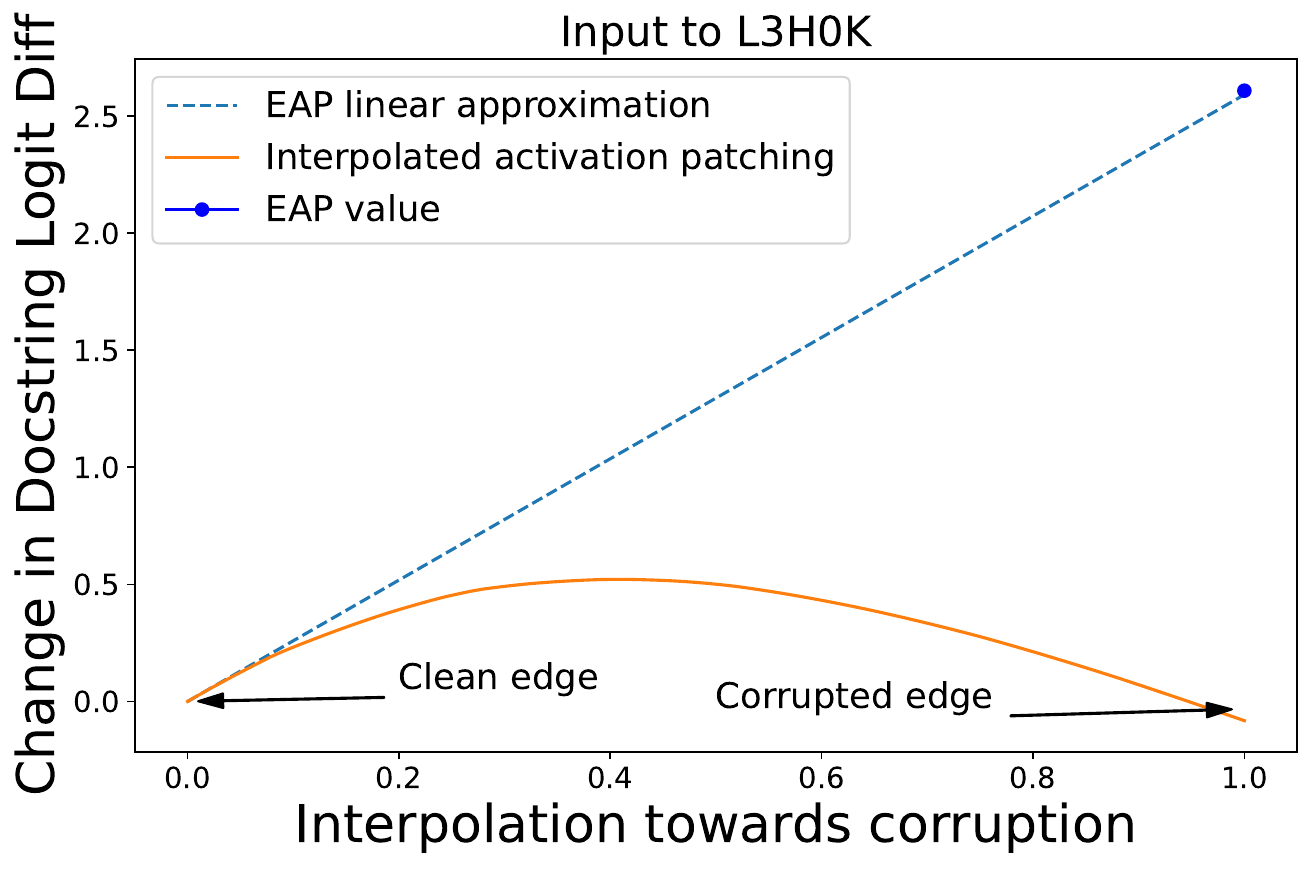}
    \end{subfigure}
    \hfill
    \begin{subfigure}[t]{0.49\textwidth}
        \centering
        \includegraphics[width=\textwidth]{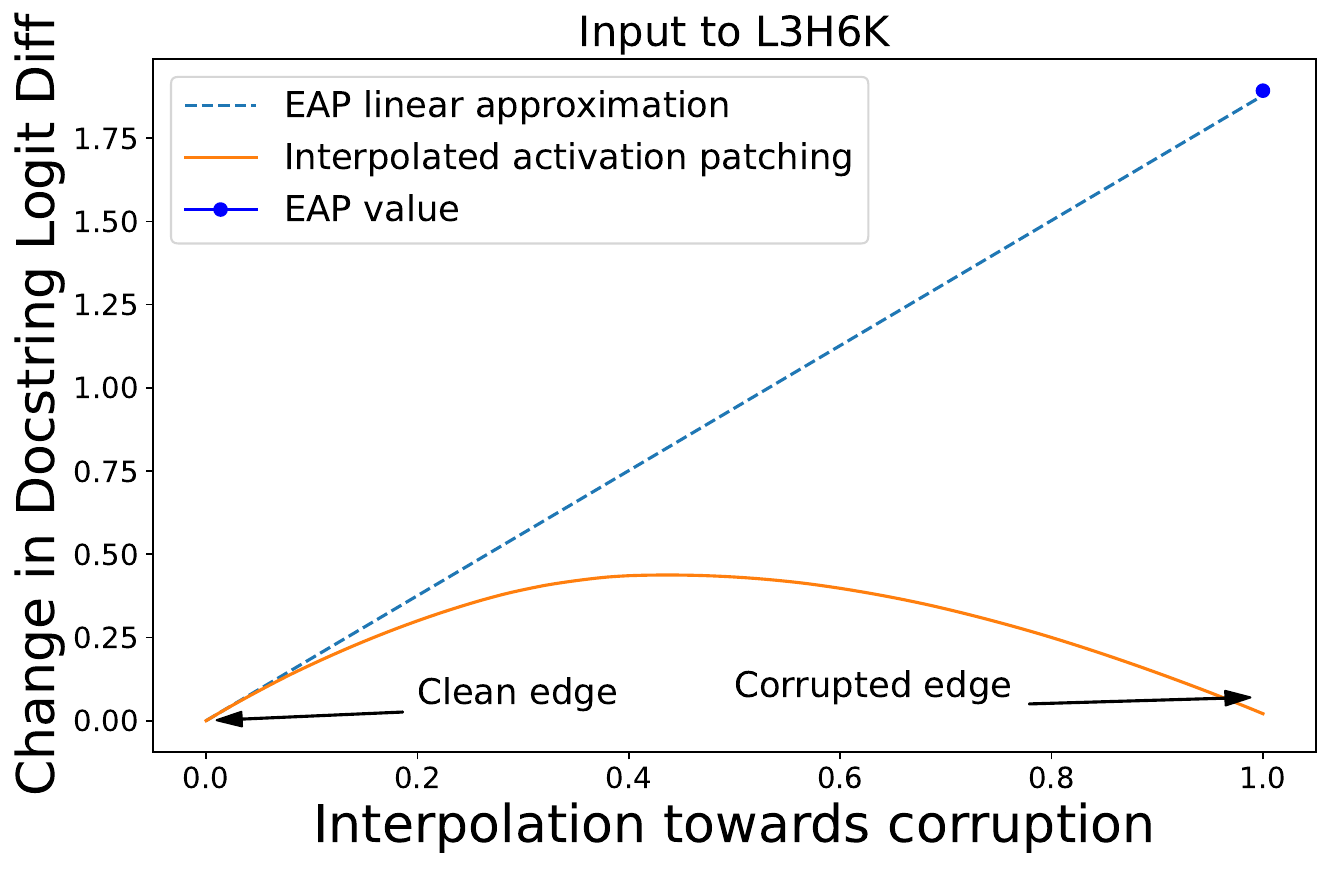}
        \label{fig:study-linear-approx2}
    \end{subfigure}
    \caption{Visualizing Edge Attribution Patching in two further cases where the concave activation patching curve means the linear fit is poor.}
    \label{fig:my-app}
\end{figure}
\newpage
\section{Edges Roles in IOI}
\label{app:edge_roles}

We further explore the attribution scores for the IOI circuit. The IOI circuit is comprised of different attention head classes such as Induction heads, S-Inhibition heads, etc. \cite{wang2023interpretability}. Figure \ref{fig:job_attr_dist_ioi} shows the distributions of scores stratified by the roles of the edges. The edge roles are defined according to the role of their origin node. While edge roles such as Previous Token, Duplicate Token, Induction, and S-Inhibition edges have attribution scores centered around zero, we see a bias in edge scores given to name mover and negative name mover edges. As the name mover edges are directly responsible for the model outputting the indirect object, the attribution scores are largely negative since ablating these edges removes the model's ability to output the indirect object, lowering the logit difference. Similarly, the negative name movers have attribution scores that are largely positive since ablating these edges improves the logit difference. This matches the intuitive function of the edges.

\begin{figure}[!h]
    \centering
    \includegraphics[width=\textwidth]{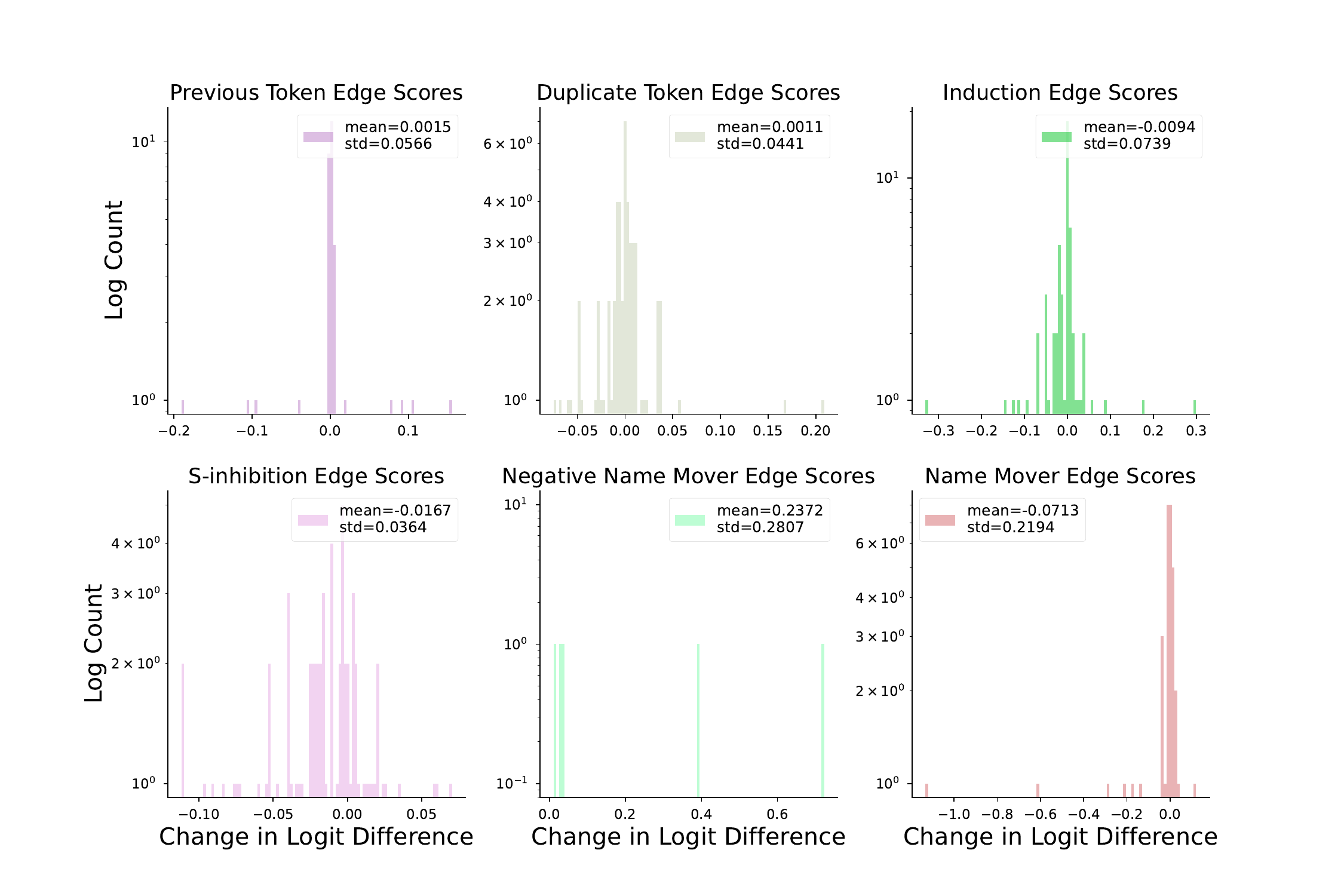}
    \caption{Distribution of Attribution Scores for each Edge Role in the IOI Task.}
    \label{fig:job_attr_dist_ioi}
\end{figure}

\section{Only one backwards pass is required for EAP}
\label{app:only_one_backwards_pass}

Note: it may be easier to understand our implementation \url{https://github.com/Aaquib111/acdcpp/blob/main/utils/prune\_utils.py#L249} rather than read this explanation. Alternatively, this derivation uses essentially the same arguments as \citet{neelattribution}\footnote{Specifically, this section: \url{https://www.neelnanda.io/mechanistic-interpretability/attribution-patching\#how-to-think-about-activation-patching=:~:text=axes\%20of\%20variation.-,Path\%20patching,-The\%20core\%20intuition}} though with an updated codebase.

There are only two types of edges iterated over in ACDC: i) residual edges where the result is added at its endpoint, and ii) edges between the residual stream and the query, key and value calculations.\footnote{It may be worth looking at some ACDC outputs from \citep{conmy2023automated}. See \url{https://colab.research.google.com/github/ArthurConmy/Automatic-Circuit-Discovery/blob/main/notebooks/colabs/ACDC\_Implementation\_Demo.ipynb} for an explanation of this design choice.} Clearly for all edges like ii) we can compute the gradient terms in \Cref{eqn:attribution_patching} in one backwards pass. 

Interestingly, for all $\Delta_e L$ terms where $e$ is a type i) edge (i.e added at the endpoint), we only need calculate the gradient with respect to the endpoint of the edge! For example, suppose we're calculating the effect of L0H0 on L1H0Q. If we represent the input to L1H0Q as a node $V$ in the computational graph then

\begin{equation}
    \frac{\partial}{\partial e_{\text{clean}}} L(\xclean |\ \text{do}(E=e_{\text{clean}})) = \frac{\partial}{\partial v_{\text{clean}}} L(\xclean |\ \text{do}(V=v_{\text{clean}}))
\end{equation}

due to how $V$ is just the sum of all the edges entering $V$. This allows efficient calculation of all the $\Delta_e L$ values since gradients with respect to nodes in computational graphs are calculated by default in backwards passes.
\end{document}

